\documentclass[runningheads]{llncs}
\usepackage[T1]{fontenc}
\usepackage{graphicx}
\begin{document}
\title{Talking to Robots: A Practical Examination of Speech Foundation Models for HRI Applications}
\titlerunning{A Practical Examination of Speech Foundation Models for HRI Applications}
\author{Theresa Pekarek Rosin \and Julia Gachot \and Henri-Leon Kordt \and Matthias Kerzel \and Stefan Wermter}
%\author{Anonymous submission}
\authorrunning{T. Pekarek Rosin et al.}
\institute{Knowledge Technology, Department of Informatics, University of Hamburg, Vogt-Koelln-Str. 30, 22527 Hamburg, Germany\\
% \email{theresa.pekarek-rosin@uni-hamburg.de}\\
\url{www.knowledge-technology.info}}
\maketitle              % typeset the header of the contribution
\begin{abstract}
    Automatic Speech Recognition (ASR) systems in real-world settings need to handle imperfect audio, often degraded by hardware limitations or environmental noise, while accommodating diverse user groups. In human–robot interaction (HRI), these challenges intersect to create a uniquely challenging recognition environment. We evaluate four state-of-the-art ASR systems on eight publicly available datasets that capture six dimensions of difficulty: domain-specific, accented, noisy, age-variant, impaired, and spontaneous speech. 
    Our analysis demonstrates significant variations in performance, hallucination tendencies, and inherent biases, despite similar scores on standard benchmarks. These limitations have serious implications for HRI, where recognition errors can interfere with task performance, user trust, and safety.

\end{abstract}
\section{Introduction}

Automatic Speech Recognition (ASR) has advanced significantly in recent years, promising zero-shot performance on most domains~\cite{Radford2022}. Yet, in real-world applications such as human–robot interaction (HRI), performance often falls short of expectations, due to corrupted audio data and diverse user groups. Training data typically covers only a narrow range of speech and noise variations, causing accuracy to drop whenever input deviates from this norm \cite{FENG2024101567}.
This contrasts with the fact that HRI research has always aimed to include people from all demographics~\cite{Asgharian2022,Raptopulou2021}, and such gaps disproportionately affect users with disabilities \cite{mujtaba-etal-2024-lost}, age-related disfluencies \cite{FENG2024101567}, accents or dialects \cite{Dhanjal2023}, and those in noisy, out-of-domain settings \cite{Vinnikov2024}.
However, for non-expert users interacting with social robots in everyday environments, a robust ASR system is essential.

In this paper, we evaluate state-of-the-art (SOTA) ASR models along the following diverse dimensions of difficulty: domain-specific, accented, noisy, age-variant, i.e., from children or elderly speakers, impaired or disordered, and spontaneous speech.
We highlight performance disparities, hallucination tendencies, and inherent biases to support informed model selection for real-world applications. These findings are particularly relevant for HRI, where ASR limitations can directly impact task effectiveness, user trust, and interaction safety.

\section{Evaluation}
\label{sec:datasets}
We select eight datasets based on their public accessibility, availability of transcriptions, and relevance in evaluating the six dimensions of difficulty:
\begin{itemize}
    \item \textbf{Cleft Dataset}~\cite{Cleland2019TheIO} (age-variant, impaired): audio from children with a cleft lip and palate, including syllables, short words, and sentences from speech therapy sessions. 
    \item \textbf{Common Voice 11.0}~\cite{Ardila2019} (domain-specific, accented, noisy, age-variant): large-scale crowd-sourced recordings of sentences from Wikipedia and classic literature.
    \item \textbf{Pitt Corpus}~\cite{pitt_corpus} (noisy, age-variant, impaired, spontaneous): spontaneous speech from storytelling, recall, and fluency assessment tasks with dementia patients and controls.
    \item \textbf{L2-ARCTIC}~\cite{zhao2018l2arctic} (domain-specific, accented, spontaneous): non-native English speakers reading CMU Arctic\footnote{\url{http://www.festvox.org/cmu\_arctic/}} and literary sentences, plus storytelling.
    \item \textbf{Nigerian English Dataset}\footnote{\url{https://openslr.org/70/}} (domain-specific, accented): sentences by Nigerian English speakers with a high number of named entities (places, people).
    \item \textbf{Speech Accent Archive}\footnote{\url{https://accent.gmu.edu}} (accented): accented speech by native speakers of diverse languages reading a phoneme-rich English paragraph.
    \item \textbf{TED-LIUM v2}~\cite{tedlium2} (domain-specific, spontaneous): TED talks on varied topics by a large number of speakers.
    \item \textbf{TORGO}~\cite{torgo} (age-variant, impaired): speech from individuals with cerebral palsy or amyotrophic lateral sclerosis, reading words and sentences from a phoneme-rich paragraph.
\end{itemize}
We prepare these datasets by converting all audio files to 16 kHz WAV format, removing unintelligible or unlabeled files, and splitting larger files sentence-wise with existing or inferred timestamps. 

Based on their widespread use and strong benchmark performance, we evaluate four state-of-the-art ASR models: \textbf{Whisper-large v3}~\cite{Radford2022}, \textbf{CrisperWhisper}~\cite{zusag24_interspeech}, \textbf{Canary-1B}~\cite{puvvada24_interspeech}, and \textbf{Parakeet-TDT-1.1B}~\cite{galvez24_interspeech}.
Model performance is measured on normalized transcripts using word error rate (WER), which quantifies incorrectly recognized words in a sentence. When models over-generate (e.g., hallucinate) tokens, it can exceed 100\%. We also analyze performance for each dimension of difficulty, averaging results across all datasets in each category. For accuracy, we exclude control speakers from the impaired category, scripted speech from the spontaneous category, and middle-aged and young adult speakers from the age-variant category.

\section{Results}

\begin{table*}[!tb]
  \caption{The WER ($\downarrow$) percentage for the ASR models on the datasets. *: The dataset is included in the training data.}
  \label{tab:wer}
  \centering  
  \begin{tabular}{l@{\hskip 0.05in}|@{\hskip 0.05in}c@{\hskip 0.05in}|@{\hskip 0.05in}c@{\hskip 0.05in}|@{\hskip 0.05in}c@{\hskip 0.05in}|@{\hskip 0.05in}c@{\hskip 0.05in}|@{\hskip 0.05in}c@{\hskip 0.05in}|@{\hskip 0.05in}c@{\hskip 0.05in}|@{\hskip 0.05in}c@{\hskip 0.05in}|@{\hskip 0.05in}c@{\hskip 0.05in}}
                     & Cleft & CV11 & Pitt & L2-A & NED & SAA & TED & TORGO \\ \hline
    Whisper-large v3& 297.08& 9.11& 32.96& 17.90& 9.55& 5.42& 10.59& 25.99\\
    CrisperWhisper    & 251.06& $^*$4.63& 57.78& 11.07& \textbf{6.64}& 6.09& 7.56& 31.88\\
    Canary-1B      & 667.74& $^*$5.26& \textbf{32.88}& \textbf{9.42}& 8.30& \textbf{2.94}& \textbf{6.26}& 30.62\\
    Parakeet-1.1B    & \textbf{169.74}& $^*$\textbf{3.15}& 37.12& 9.56& 7.48& 3.24& 6.56 &\textbf{25.93} \\
    
  \end{tabular}
  
\end{table*}

\begin{table*}[!tb]
  \caption{The WER ($\downarrow$) percentage for all models across the dimensions of difficulty.}
  \label{tab:wer_dod}
  \centering  
  \begin{tabular}{l@{\hskip 0.05in}|@{\hskip 0.05in}c@{\hskip 0.05in}|@{\hskip 0.05in}c@{\hskip 0.05in}|@{\hskip 0.05in}c@{\hskip 0.05in}|@{\hskip 0.05in}c@{\hskip 0.05in}|@{\hskip 0.05in}c@{\hskip 0.05in}|@{\hskip 0.05in}c}
                     & Domain-Sp. & Accented & Noisy & Age-Var. & Imp. & Spont. \\ \hline
    Whisper-large v3 & 11.79& 10.50& 21.04& 73.74& 130.88& 18.70\\
    CrisperWhisper   & 7.47& 7.11& 31.20& 69.86& 123.72&  24.36\\
    Canary-1B        & 7.31& 6.48& \textbf{19.07}& 148.03& 258.48&   \textbf{17.57}\\
    Parakeet-1.1B    & \textbf{6.69}& \textbf{5.86}& 20.13& \textbf{47.71}& \textbf{86.09}&   19.50\\
    
  \end{tabular}
  
\end{table*}

As shown in Table~\ref{tab:wer}, performance varies widely across datasets and with the number of dimensions involved.
The highest-WER cases are typically caused by hallucinations, especially for single-word ground truths (Cleft Dataset) and disordered speech (Pitt Corpus). This is particularly problematic for HRI, where speech commands are short, and older, potentially impaired, speakers are a key user group~\cite{Asgharian2022}.
In some instances, models misidentify the language, most commonly Arabic or Korean, producing random character sequences.
Parakeet-TDT-1.1B appears less prone to this issue, likely due to its Transducer decoder’s character-based decoding, which limits hallucinations compared to subword-based methods.
Accented speech is generally well recognized when scripted (CV 11.0, SAA), but lower scores on NED and L2-ARCTIC indicate that domain-specific vocabulary and spontaneous speech also contribute to the overall difficulty of a sample.

Overall, as shown in Table~\ref{tab:wer_dod}, models perform well in isolated dimensions (e.g., spontaneous speech) but their performance degrades when multiple variations, such as background noise or speech impairments, coincide. This suggests an additive penalty as speech diverges from the training data and audio quality deteriorates.
While accented and domain-specific speech often reach acceptable WERs (<10\%), spontaneous, disordered, age-variant, and noisy speech remain challenging, especially in combination. On average, Parakeet-TDT-1.1B outperforms the other models by 10–50\%, with Canary-1B surpassing it only slightly for spontaneous and noisy speech. Based on these results, either Parakeet-TDT-1.1B or Canary-1B is recommended for HRI applications, but real-time performance must also be rigorously assessed to ensure practical deployment.

\begin{credits}
\subsubsection{\ackname} 
The authors gratefully acknowledge funding from Horizon Europe under the MSCA grant agreement No 101072488 (TRAIL).

\subsubsection{\discintname} The authors have no competing interests to declare that are
relevant to the content of this article.
\end{credits}

\bibliographystyle{splncs04}
\bibliography{mybib}

\begin{thebibliography}{10}
\providecommand{\url}[1]{\texttt{#1}}
\providecommand{\urlprefix}{URL }
\providecommand{\doi}[1]{https://doi.org/#1}

\bibitem{Ardila2019}
Ardila, R., Branson, M., Davis, K., et~al.: {Common Voice: A Massively-Multilingual Speech Corpus}. In: Proc. LREC 2020. pp. 4218--4222. Marseille, France (2020)

\bibitem{Asgharian2022}
Asgharian, P., Panchea, A.M., Ferland, F.: A review on the use of mobile service robots in elderly care. Robotics  \textbf{11}(6) (2022)

\bibitem{pitt_corpus}
Becker, J.T., Boiler, F., Lopez, O.L., et~al.: {The Natural History of Alzheimer's Disease: Description of Study Cohort and Accuracy of Diagnosis}. Archives of Neurology  \textbf{51}(6),  585--594 (1994)

\bibitem{Cleland2019TheIO}
Cleland, J., Lloyd, S., Campbell, L., et~al.: {The Impact of Real-Time Articulatory Information on Phonetic Transcription: Ultrasound-Aided Transcription in Cleft Lip and Palate Speech}. Folia Phoniatrica et Logopaedica  \textbf{72},  120--130 (2019)

\bibitem{Dhanjal2023}
Dhanjal, A.S., Singh, W.: A comprehensive survey on automatic speech recognition using neural networks. Multimedia Tools and Applications  \textbf{83}(8),  23367--23412 (2023)

\bibitem{FENG2024101567}
Feng, S., Halpern, B.M., Kudina, O., et~al.: Towards inclusive automatic speech recognition. Computer Speech and Language  \textbf{84}(C),  101567 (2024)

\bibitem{galvez24_interspeech}
Galvez, D., Bataev, V., Xu, H., et~al.: {Speed of Light Exact Greedy Decoding for RNN-T Speech Recognition Models on GPU}. In: Proc. Interspeech 2024. pp. 277--281. Kos, Greece (2024)

\bibitem{mujtaba-etal-2024-lost}
Mujtaba, D., Mahapatra, N., Arney, M., et~al.: {Lost in Transcription: Identifying and Quantifying the Accuracy Biases of Automatic Speech Recognition Systems Against Disfluent Speech}. In: Proc. NAACL-HLT 2024. pp. 4795--4809. Mexico City, Mexico (2024)

\bibitem{puvvada24_interspeech}
Puvvada, K.C., Żelasko, P., Huang, H., et~al.: {Less is More: Accurate Speech Recognition \& Translation without Web-Scale Data}. In: Proc. Interspeech 2024. pp. 3964--3968. Kos, Greece (2024)

\bibitem{Radford2022}
Radford, A., Kim, J.W., Xu, T., et~al.: {Robust Speech Recognition via Large-Scale Weak Supervision}. In: Proc. ICML 2023. pp. 28492--28518. {Honolulu, Hawaii, USA} (2023)

\bibitem{Raptopulou2021}
Raptopoulou, A., Komnidis, A., Bamidis, P.D., et~al.: Human–robot interaction for social skill development in children with asd: A literature review. Healthcare Technology Letters  \textbf{8}(4),  90--96 (2021)

\bibitem{tedlium2}
Rousseau, A., Del{\'e}glise, P., Est{\`e}ve, Y.: {Enhancing the {TED}-{LIUM} Corpus with Selected Data for Language Modeling and More {TED} Talks}. In: Proc. LREC 2014. pp. 3935--3939. Reykjavik, Iceland (2014)

\bibitem{torgo}
Rudzicz, F., Namasivayam, A.K., Wolff, T.: {The TORGO database of acoustic and articulatory speech from speakers with dysarthria}. Language Resources and Evaluation  \textbf{46},  523 -- 541 (2011)

\bibitem{Vinnikov2024}
Vinnikov, A., Ivry, A., Hurvitz, A., et~al.: {NOTSOFAR-1 Challenge: New Datasets, Baseline, and Tasks for Distant Meeting Transcription}. In: Proc. {Interspeech} 2024. pp. 5003--5007. Kos, Greece (2024)

\bibitem{zhao2018l2arctic}
{Zhao}, G., {Sonsaat}, S., {Silpachai}, A., et~al.: {L2-ARCTIC: A Non-native English Speech Corpus 2018}. In: Proc. Interspeech 2018. pp. 2783--2787. Hyderabad, India (2018)

\bibitem{zusag24_interspeech}
Zusag, M., Wagner, L., Thallinger, B.: {CrisperWhisper: Accurate Timestamps on Verbatim Speech Transcriptions}. In: Proc. Interspeech 2024. pp. 1265--1269. Kos, Greece (2024)

\end{thebibliography}

\end{document}